\documentclass[11pt]{article}

\usepackage[margin=1in]{geometry}
\usepackage{booktabs}
\usepackage{amsmath}
\usepackage{microtype}
\usepackage[round]{natbib}
\usepackage[hidelinks]{hyperref}
\usepackage{newtxtext,newtxmath}

\newcommand{\code}[1]{\texttt{#1}}

\title{Committed Before Reasoning: Behavioral Reproduction and Preliminary
Activation-Level Evidence of Answer Pre-Commitment in an Open-Weight LLM}
\author{Heejin Jo}
\date{July 2026}

\begin{document}
\maketitle

\begin{abstract}
Chat models sometimes commit to an answer and then produce reasoning that
justifies the commitment rather than deriving it --- even when the committed
answer contradicts a premise of the task. We study a minimal probe of this
failure: ``I want to wash my car. The car wash is 100 meters away. Should I
walk or drive?'' The only correct answer is \emph{drive}, because the car
must be present at the car wash; models overwhelmingly recommend walking. We
make three contributions. \textbf{(1) Behavioral reproduction.} On Qwen3-8B,
across five system-prompt conditions (210 rollouts), the wrong commitment
occurs in 85--100\% of sampled rollouts per condition and in 100\% of greedy
rollouts, in both thinking and non-thinking modes; extended chain-of-thought
does not repair it (walk-rate 85--100\% with a 4{,}096-token thinking
budget). \textbf{(2) Preliminary activation-level evidence.} Using a pretrained,
\emph{training-free} activation oracle (no task-specific probe training),
we probe hidden states at positions
\emph{before} the answer text is emitted. Oracle read-outs of ``walk'' at
pre-commit positions exceed a neutral-context baseline (68\% vs.\ 17\%;
walk-committing rollouts $p=.005$, drive-committing rollouts $p=.005$,
Fisher exact) --- and, notably, rollouts that eventually answer \emph{drive}
also read as walk-leaning before commitment (5/6). Because the oracle's
default answer on unrelated content is ``drive'' (83\%), these walk
read-outs are not explained by lexical bias; a lexical stratification
further shows they are not text recovery --- spans containing ``drive''
still read out walk, and in balanced lexical fields naming both options,
per-rollout walk-majorities dominate a per-prompt neutral baseline (15/22
vs.\ 1/8, $p=.01$; drive-committing rollouts 6/6, $p=.002$).
Sample sizes are small and the within-rollout positional gradient is not
significant ($p=.34$); we frame these results as preliminary.
\textbf{(3) Methodological.} The same oracle, activations, and positions
succeed or fail depending almost entirely on question wording: an open
question (``What answer is this model going to give?'') fails a positive
control (2/16) that a closed alternative (``Is the model going to say walk
or drive?'') passes (11/16). Current activation oracles are usable, but
brittle in ways that standard practice does not yet document.
\end{abstract}

\section{Introduction}

A model that answers before it reasons is not merely being terse. When the
answer is fixed first, subsequent reasoning tends to be \emph{advocacy} ---
locally coherent argumentation for the fixed answer --- rather than
derivation. The failure is invisible exactly when it matters: the reasoning
reads as diligent, and only a premise-level check exposes that the
conclusion was never derived from it.

We study a deliberately minimal instance. The car-wash question (``The car
wash is 100 meters away. Should I walk or drive?'') has a single premise
that decides it: \emph{the car must be at the car wash for the car to be
washed}. Walking optimizes a nearby but wrong objective (short trips are
better on foot). The question was originally observed to elicit confident
``walk'' answers, followed by fluent justification, from Claude Sonnet 4.5
(\code{claude-sonnet-4-5-20250929}); this paper asks (i) whether the
phenomenon reproduces on an open-weight model at measurable rates, and
(ii) whether a \emph{training-free} activation oracle can read the
commitment before the answer text appears. The second question is
deliberately narrow: a fast-growing line of work has now established,
with supervised linear probes, that answers are often encoded in
activations before reasoning text is generated
\citep{cox,boppana,mirtaheri,scalena}. What remains open is whether
training-free natural-language read-out --- which prior work found
unreliable on safety-relevant tasks --- can see the same thing, and
whether it holds in a setting where the commitment is self-generated by
surface heuristics rather than induced by a hint.

Our direction is the reverse of hint-injection studies of unfaithful
chain-of-thought (e.g., attribution-graph analyses in which a planted answer
bends the reasoning): there, reasoning is corrupted toward a given answer;
here, the reasoning is often \emph{sound in isolation} --- the model
enumerates factors, weighs them, sometimes even touches the critical premise
--- and the answer ignores it. Qualitatively (Section~\ref{sec:qual}), the
model walks up to the door (``if the car is already parked near the car
wash, maybe\ldots'') and does not open it.

\section{Behavioral study}

\subsection{Setup}

\paragraph{Model and decoding.} Qwen3-8B \citep{qwen3}, run locally on
Apple silicon (MPS) in bf16 --- the 8-bit path used by the oracle
reference setup is CUDA-only (bitsandbytes), and bf16 is the standard
full-quality precision that fits the model in unified memory. Two decoding
settings per condition and thinking mode: greedy (deterministic; one
rollout as a reference point) and temperature-0.7 sampling ($n=20$,
matching the original experiment's per-condition $n$) to observe the
commitment distribution. Thinking mode toggled via the chat template
(\code{enable\_thinking}).

\paragraph{Conditions.} Five system-prompt conditions taken verbatim from
the original experiment: \code{A\_bare} (no system prompt),
\code{B\_role\_only}, \code{C\_role\_star} (role + mandatory STAR answer
structure), \code{D\_role\_profile} (role + user profile stating the car is
parked in the driveway), \code{E\_full\_stack} (B + STAR + profile). Per
condition and thinking mode: 1 greedy + 20 sampled rollouts (210 total),
plus a follow-up challenge turn (``How will I get my car washed if I am
walking?'') for recovery analysis.

\paragraph{Scoring.} The primary metric is \code{committed\_wrong}: the
first recommendation in the speaker's own voice is \emph{walk}. We
initially scored with a rule-based (regex) scorer, audited it
adversarially, and rejected it when it failed a fresh 32-case holdout
(62.5\% case-level disagreement with human labels; almost all errors were
missed commitments). We replaced it with an LLM judge (Qwen3-8B, greedy,
rubric prompt, JSON output), adopted only after passing a pre-registered gate (acceptance threshold
$\geq$95\%, fixed before scoring): 96.9\% (62/64) agreement with human labels on synthetic audit cases
\emph{not used to tune the judge prompt}, with label-level determinism
verified (8/8 identical on double-scoring). A stratified manual gate over real rollouts --- mandated by the
experiment spec at 15--20 cases; we used 20 (20/20 correct labels,
hand-checked) --- closed the loop. A
secondary construct (\code{answer\_before\_reasoning}) failed validation
(82.8\%) and is excluded from all claims. Appendix~\ref{app:audit}
summarizes the audit trail.

\subsection{Results: the wrong commitment is near-deterministic}
\label{sec:behavior}

\begin{table}[ht]
\centering
\begin{tabular}{lcc}
\toprule
Condition & thinking off & thinking on \\
\midrule
\code{A\_bare}         & 85\% (100\%)  & 85\% (100\%) \\
\code{B\_role\_only}   & 90\% (100\%)  & 90\% (100\%) \\
\code{C\_role\_star}   & 100\% (100\%) & 100\% (100\%) \\
\code{D\_role\_profile}& 100\% (100\%) & 85\% (100\%) \\
\code{E\_full\_stack}  & 100\% (100\%) & 85\% (100\%) \\
\bottomrule
\end{tabular}
\caption{Wrong-commitment rates (\code{judge\_committed\_wrong}, sampled
$n=20$ per cell; greedy in parentheses).}
\label{tab:rates}
\end{table}

Across all 210 rollouts, committed answers split walk 194 / drive 10 /
none 6: 95\% of all commitments are wrong. The failure is not an occasional
sampling accident; at temperature 0.7 it is the modal behavior everywhere,
and greedy decoding produces it in 10/10 cells.

Three structural observations:
\begin{enumerate}
\item \textbf{Structured-answer instructions maximize the failure.}
\code{C\_role\_star} (mandatory STAR format) is at 100\% in both thinking
modes. A format that forces an early ``Action:'' slot appears to lock the
commitment in.
\item \textbf{Extended thinking does not repair it.} With a 4{,}096-token
thinking budget (Section~\ref{sec:trunc}), thinking-mode walk rates remain
85--100\%. Thinking contributes ${\sim}5$ percentage points of rescue: 9 of
the 10 correct (drive) commitments in the dataset occur in thinking mode.
\item \textbf{The profile does not help.} \code{D\_role\_profile}
explicitly states the car is parked in the user's driveway; the
wrong-commitment rate is 85--100\% regardless.
\end{enumerate}

\subsection{A truncation artifact that almost reversed a conclusion}
\label{sec:trunc}

Our first full run used a 1{,}024-token generation budget. Thinking-mode
rollouts showed apparently lower wrong-commitment rates (45--95\% across
conditions, vs.\ 85--100\% after the fix), which would have supported
``thinking mitigates the failure.'' Inspection of the 35 null-commitment
thinking rollouts showed 31 were \code{<think>} blocks truncated
mid-thought by the budget --- no visible answer existed to score (the
remaining 4 were genuine non-commitments). After regenerating the entire thinking arm at
4{,}096 tokens (0/210 truncated), the mitigation disappeared. We report
this because aggregate-only pipelines would have shipped the wrong
conclusion; the correction came from reading the raw texts behind an
anomalous label cluster.

\subsection{Qualitative anatomy of one rollout}
\label{sec:qual}

In a representative \code{C\_role\_star} thinking rollout, the think block
frames the task as a preference choice --- enumerating walking speed,
weather, and effort --- brushes against the decisive fact exactly once
(``if the car is already parked near the car wash, maybe\ldots'') and
moves on without resolving where the car actually is. The visible answer
then fills the STAR template: ``\textbf{Action}: Walk to the car wash.''
The reasoning is diligent and internally consistent; it is also
premise-blind: the one question that decides the task (\emph{how does the
car get there?}) is raised obliquely and never answered. This is the shape
of the phenomenon: not corrupted reasoning, but an answer that reasoning
never actually authorized.

\section{Probing activations before the commitment (preliminary)}

If the model is committed to ``walk'' before emitting it, the commitment
may be readable from hidden states at positions preceding the answer text.
We test this with a pretrained activation oracle (LatentQA-style: an LLM
fine-tuned to answer natural-language questions about injected activations
\citep{latentqa}), using the public Qwen3-8B oracle checkpoint
\citep{oracles}. Activations are collected by teacher-forced prefill over
the exact generation-time sequence (prompt + generated text), at layer 18
of 36 --- 50\% depth, the oracle checkpoint's training and demo default.
We did not sweep layers (see Limitations); the per-question positive
control below partially de-risks this for the question we adopt.

\paragraph{Design.} 22 rollouts from the behavioral study.
Walk-committing: the first four eligible per cell in dataset order, from
\code{A\_bare} (the bare condition) and \code{C\_role\_star} (the
strongest-failure condition) $\times$ thinking on/off. Eligibility = a
localizable commitment position: we anchor the position grid on the
rejected regex scorer's commitment span, used only where it agreed with
the judge label (the span supplies a \emph{position}, not a label), plus
the Section~\ref{sec:qual} exemplar, whose span we hand-localized.
Drive-committing: the dataset contains 10; 6 satisfied the same
localizability rule (in the other 4 the regex found no span to anchor on).
Selection is deterministic given the data but not random, and the
localization rule inherits the regex scorer's blind spots; with samples
this small we treat every statistic in this section as preliminary. Six probe sites per rollout: assistant start
(P0), 25/50/75\% of the span to the commitment token (P1--P3), immediately
before the commitment token (P4), and a span starting at the commitment
token (P5 = [commit, +8 tokens], positive control). Oracle decoding is
greedy throughout. P0--P4 windows never touch the commitment tokens, so a
correct read-out at those sites cannot come from the answer text itself
(text-inversion control).

\subsection{The oracle is extremely sensitive to question wording}
\label{sec:wording}

Our first three configurations --- single-token probes, 5-token majority
voting, and segment (multi-position) injection --- all failed the positive
control with the open question ``What answer is this model going to
give?'' (P5 hit rates 3/16, 3/16, 2/16). The failure was consistent, not
noisy: under majority voting, 10/16 rollouts produced 4--5 ``no answer''
votes out of 5 at P5. Our pre-set recovery criterion for the positive
control, fixed before the question comparison, was roughly 10/16.

Holding activations, positions, and injection mode fixed and changing only
the question:

\begin{table}[ht]
\centering
\begin{tabular}{lc}
\toprule
Oracle question (P5, walk group) & hits \\
\midrule
``What answer is this model going to give?'' & 2/16 \\
``What is the model recommending?'' & 7/16 \\
``Is the model going to say walk or drive?'' & \textbf{11/16} \\
\bottomrule
\end{tabular}
\caption{Positive-control recovery under question rewording (segment
injection, spans including the emitted commitment tokens).}
\label{tab:wording}
\end{table}

A closed question naming the answer space recovers the positive control
that open questions fail. We conjecture open answer-prediction is outside
the oracle's LatentQA training distribution while binary classification is
close to it. Practical implication: a negative oracle result cannot be
interpreted without a positive control \emph{per question wording} --- a
constraint we have not seen documented in current oracle practice, and an
independent, quantified confirmation that current activation oracles are
hard to use off-distribution.

\subsection{Neutral-context baseline: the oracle's default is ``drive''}
\label{sec:neutral}

Forced-choice questions make the oracle answer something even on empty
evidence, so read-out rates are only interpretable against the oracle's
intrinsic preference. On activations from 8 unrelated prompts (philosophy,
arithmetic, recipes, code; 24 probes), the oracle answers ``drive'' 19/23
(83\%) and ``walk'' 4/23 (17\%); one probe mentioned both options and is
excluded from the decisive-read-out denominator. We pre-registered
interpretation bands before running the control (40--60\% walk share would
support genuine signal; $\geq$70\% would indicate walk bias); the observed
17\% fell outside both bands, in the direction opposite the bias concern. The oracle's default under this question is
\emph{drive}; consequently ``walk'' read-outs carry information and
``drive'' read-outs are weak evidence. (This asymmetry cuts both ways: it
strengthens walk read-outs below, and it means our positive control stands
only on the walk side --- drive-committing rollouts' P5 read-outs, 3/6,
are uninformative.)

\subsection{Pre-commit read-outs exceed the baseline --- including in
rollouts that end up correct}
\label{sec:precommit}

\begin{table}[ht]
\centering
\begin{tabular}{lcc}
\toprule
Group & P4 walk read-outs & vs.\ baseline \\
\midrule
walk-committing ($n=16$) & 10/16 (63\%) & $p = .005$ \\
drive-committing ($n=6$) & 5/6 (83\%) & $p = .005$ \\
combined ($n=22$) & 15/22 (68\%) & $p = .0007$ \\
\bottomrule
\end{tabular}
\caption{Walk read-out rates at the pre-commit site P4 (commitment text
not yet emitted), against the 17\% neutral baseline (Fisher exact,
one-sided).}
\label{tab:precommit}
\end{table}

Two readings, stated with their limits:
\begin{enumerate}
\item \textbf{The wrong commitment is readable before it is written ---
by a training-free oracle.} That pre-CoT answer encoding exists is
established with supervised probes \citep{cox}; the increment here is
that a natural-language oracle with no task-specific training reads it
too: in walk-committing rollouts, an oracle whose default answer is
``drive'' reads ``walk'' from pre-commit activations at $3.6\times$ the
neutral rate (one-sided tests throughout; direction fixed in advance).
\item \textbf{Rollouts that eventually answer correctly also start
walk-leaning.} 5/6 drive-committing rollouts read as ``walk'' at P4.
Since 9/10 drive commitments arise in thinking mode after long
deliberation (Section~\ref{sec:behavior}), this is consistent with a
two-stage picture: \emph{walk is the default internal state;
occasionally, extended reasoning overrides it late.} The behavioral and
activation-level views agree on where the default points.
\end{enumerate}

\paragraph{What we do not claim.} The within-rollout positional gradient
(P0 6/16 $\rightarrow$ P1 8 $\rightarrow$ P2 9 $\rightarrow$ P3 12
$\rightarrow$ P4 10 $\rightarrow$ P5 11, walk group) trends upward but is
not significant (paired sign test P0 vs.\ P4: 7 improved / 3 worsened,
$p = .34$). P0 activations already contain the prompt (the question is
visible to the model), so P0 is not a zero-information site --- elevation
\emph{at} P0 relative to neutral is expected and observed (6/16 vs.\
17\%). With $n=16$ we cannot distinguish ``commitment strengthens toward
the answer'' from ``commitment is present throughout at roughly constant
readability.'' The drive-group result rests on $n=6$ and is preliminary by
any standard. Individual-rollout trajectories are noisy (the
Section~\ref{sec:qual} rollout reads walk at P2, drive at P4, walk at P5).

\subsection{Lexical stratification: the read-outs are not text recovery}
\label{sec:lexical}

The remaining confound is lexical leakage: probe windows sit inside
deliberation text that frequently contains the word ``walk,'' so the
oracle might be reading nearby tokens rather than state. We stratify all
110 pre-commit probes by whether walk/drive literally occur (a) in the
injected 5-token span itself and (b) within $\pm$25 tokens of it.

\paragraph{Injected-span (strict text-inversion) test.} Walk read-outs are
\emph{not} driven by ``walk'' tokens in the injected span: spans
containing ``walk'' read out walk at 2/8, spans without it at 64/102
(63\%). Eight probes have injected spans containing ``drive''/``driving''
and not ``walk''; five of them still read out walk (e.g., span `` not need
to drive.'' $\rightarrow$ ``walk''). Recovery of injected token identities
cannot explain the results. (Word matching uses strict boundaries ---
``driveway'' does not count as ``drive.'')

\paragraph{Context-field test.} The $\pm$25-token lexical field does
influence the oracle in single-word strata: contexts containing only
``walk'' read out walk 12/13; contexts containing only ``drive'' lean
drive (3/6). These strata are small because deliberation text is saturated
with both words: 89/110 probes (81\%) have a \emph{balanced} field
containing both. Descriptively, walk read-outs dominate that stratum 52
vs.\ 20 (a 72\% walk share against the oracle's 17\% neutral default).
Because probes are clustered within rollouts (five per rollout), pooled
probe-level tests overstate independence; our primary inference is
therefore cluster-aware, comparing per-rollout majorities in the balanced
field against per-prompt majorities in the neutral control (walk-majority
prompts: 1/8). Walk-committing rollouts: 9/16 walk-majorities ($p = .051$,
marginal). Drive-committing rollouts: \textbf{6/6} walk-majorities
($p = .002$). Combined: 15/22 ($p = .010$). The drive-group cell remains
the sharpest: the lexical field names both options, the oracle's default
is drive, the rollout's own final answer is drive --- and every one of the
six rollouts reads walk on majority. Pure lexical reading predicts none of
this.

We conclude the pre-commit walk read-outs cannot be reduced to text
recovery, while acknowledging that in single-word lexical fields oracle
answers do track the field, so probes there (a minority) are individually
uninterpretable.

\section{Related work}

\paragraph{CoT faithfulness.} That stated reasoning need not reflect the
causal process behind an answer is established: biasing features silently
flip answers while the explanation never mentions them \citep{turpin},
intervention tests show models often reach the same answer with corrupted
or truncated reasoning \citep{lanham}, and reasoning models verbalize
hints they demonstrably used only 25--39\% of the time \citep{chen}. Our
phenomenon is a complement rather than an instance: in those settings the
reasoning is bent toward an externally planted answer, whereas here no
hint exists --- the reasoning is often locally sound and premise-aware,
and the commitment simply precedes and ignores it.

\paragraph{Mechanistic accounts of answer-first computation.}
Attribution-graph analyses show language models plan outputs before
emitting them (e.g., selecting a rhyme target before writing the line) and
construct post-hoc justifications in hint-driven settings
\citep{lindsey,ameisen}. Those results are single-forward-pass circuit
analyses on a closed model; we ask the coarser but complementary question
of whether a commitment is \emph{readable} from open-weight activations at
pre-emission positions, using only public tooling.

\paragraph{Pre-generation probing of decisions.} A rapidly forming
cluster establishes that answers are encoded in activations before
reasoning text, using supervised probes. \citet{cox} decode the final
answer from the residual stream at the last pre-CoT token ($>$0.9 AUC on
most tasks), show the direction is causal via steering (flipping answers
in over half of cases), and taxonomize the induced failures as
non-entailment vs.\ confabulation. \citet{boppana} show the final answer
is decodable far earlier in the CoT than a monitor can tell, and exploit
it for probe-guided early exit; \citet{scalena} locate a sharp
\emph{commitment boundary} inside reasoning chains after which further
steps are epiphenomenal. \citet{mirtaheri} detect hint-induced
\emph{motivated reasoning} from pre-generation representations, matching
or beating full-CoT monitors; \citet{esakkiraja} decode tool-calling
decisions pre-generation and flip them by steering. We treat
pre-generation answer encoding as established by this literature and do
not claim it as a contribution. Our increment is orthogonal on three
axes: (a) the read-out tool is a \emph{training-free} natural-language
oracle rather than a supervised probe --- which is what exposes the
question-wording brittleness of Section~\ref{sec:wording}, a failure
mode supervised probes cannot have; (b) the commitment is self-generated
by the task's surface heuristics rather than induced by a hint, and
reasoning that touches the decisive premise still fails to override it;
(c) we measure the behavioral failure rate jointly with the internal
signal. Triangulated with \citet{oracles}' report of substantial task
variance in oracle performance, the picture is: supervised probes read
pre-CoT commitments well; training-free oracles mostly do not ---
unless, as Section~\ref{sec:wording} shows, the question is asked
in-distribution.

\paragraph{Activation-to-language decoders.} LatentQA fine-tunes an LLM to
answer natural-language questions about injected activations
\citep{latentqa}; Activation Oracles extend this to general-purpose
activation explainers trained across diverse tasks and released for
several open models \citep{oracles}. Karvonen et al.\ report substantial
variance across downstream tasks and note calibration limitations; our
question-wording result (Section~\ref{sec:wording}) is a quantified
instance of that usability gap, and our stratified controls
(Sections~\ref{sec:neutral}--\ref{sec:lexical}) are, to our knowledge, the
first text-inversion analysis of these oracles on an answer-commitment
task.

\section{Limitations}

\begin{itemize}
\item \textbf{Single model, single task.} One 8B open-weight model, one
question. The original phenomenon was observed on a much larger closed
model; we show transfer of the behavior, not universality.
\item \textbf{Judge and target share a base model.} The LLM judge is
Qwen3-8B scoring Qwen3-8B rollouts. The 96.9\% human-agreement gate and
20/20 manual check mitigate this; raw texts are preserved so any external
judge can re-score.
\item \textbf{Probing sample sizes and clustering.} $n=16/6$ rollouts; the
positional gradient is non-significant; drive-side positive control
unavailable (oracle default). Probes are clustered within rollouts, so
pooled probe-level tests are descriptive only and inference is
cluster-aware (Section~\ref{sec:lexical}); the walk-group balanced-field
test is only marginal ($p=.051$) at rollout level. All Section~3 claims
are labeled preliminary.
\item \textbf{Single probe layer.} All probes use the oracle's default
layer (50\% depth). A layer sweep was planned but not run; results may
differ at other depths.
\item \textbf{Non-random rollout selection.} Section~3 rollouts were
selected deterministically (first-$k$ eligible per cell) under a
localizability rule that inherits the rejected regex scorer's blind spots;
4 of 10 drive-committing rollouts were excluded by it.
\item \textbf{Teacher-forced prefill.} Probed activations come from
re-encoding the generated sequence, which matches generation-time
computation for the same prefix under causal attention, but small
tokenizer boundary effects at the prompt/generation seam are possible.
\item \textbf{Local lexical context.} Addressed directly in
Section~\ref{sec:lexical}: injected-span text-inversion is refuted, and
the dominant balanced-field stratum favors walk against both the oracle
default and a lexical tie. Residual caveat: in single-word lexical fields
the oracle tracks the field, so individual probes in those (minority)
strata remain uninterpretable, and $\pm$25 tokens is one
operationalization of ``local'' among several.
\end{itemize}

\section{Discussion}

The behavioral result is sturdy: on this task, a competent open-weight
model commits to a premise-violating answer at 85--100\% rates that
survive sampling, prompt variation, an explicit contradicting profile, and
a 4{,}096-token thinking budget --- and structured-answer formats push it to the ceiling
(\code{C\_role\_star}: 100\% in both modes). The activation-level result, while preliminary, points the same way:
the default internal state reads as ``walk'' before any answer is
written, even in the rare rollouts that end up correct. The
supervised-probe literature has already established that
``answer-first'' is a readable internal condition
\citep{cox,boppana,scalena}, and that detection there can precede and
outperform reasoning monitors \citep{mirtaheri} --- supporting
pre-commitment detection, rather than post-hoc reasoning audits, as the
intervention point. Our contribution to that picture is narrower: the
condition is readable even without training a probe, in a hint-free
setting, if and only if the oracle is questioned in-distribution.

The methodological finding stands on its own: with a fixed oracle and
fixed activations, question wording alone moved a positive control from
2/16 to 11/16. Negative oracle results without per-wording positive
controls are uninterpretable, and current oracles' usable surface is
narrower than their interface suggests.

\appendix

\section{Scorer audit trail (summary)}
\label{app:audit}

Regex scorer: 6 general-rule fixes after a 32-case adversarial audit;
fresh 4-lens 32-case holdout $\rightarrow$ 62.5\% case mismatch (1 false
positive, 43/44 field errors were missed commitments) $\rightarrow$
rejected under a pre-registered $>$15\% threshold. LLM judge: rubric
frozen before validation; 62/64 on combined audit sets; the two
disagreements are pre-flagged label-ambiguous classes (implied commitment
under negation; conditional dual recommendation). Determinism: 8/8
identical labels on repeated greedy scoring (MPS). Manual gate: stratified
20 rollouts (all conditions, both thinking modes, truncation cases
included), 20/20 judge labels confirmed by hand.

\section{Reproducibility}
\label{app:repro}

All rollouts, judge outputs, probe responses, and scripts are committed:
behavioral harness (\code{stage1\_qwen\_carwash.py}, resume-capable),
judge (\code{judge\_rollouts.py}, \code{--validate} gate), AO pipeline
(\code{stage2\_ao\_\{experiment,window,segment,qab,neutral\}.py} plus
\code{stage2\_lexical\_strata.py}, vendored demo library with a 4-line
transformers-compat patch), and every statistic in the text as
data-plus-code (\code{compute\_statistics.py} $\rightarrow$
\code{stage2\_results/statistics.json}). Hardware: single Apple M-series
machine (MPS, bf16); no CUDA, quantization, or cloud GPU required.


\begin{thebibliography}{9}

\bibitem[Ameisen et al.(2025)]{ameisen}
Ameisen, E., Lindsey, J., Pearce, A., et al.\ (2025).
\emph{Circuit Tracing: Revealing Computational Graphs in Language Models.}
Transformer Circuits Thread.
\url{https://transformer-circuits.pub/2025/attribution-graphs/methods.html}

\bibitem[Boppana et al.(2026)]{boppana}
Boppana, S., Ma, A., Loeffler, M., Sarfati, R., Bigelow, E., Geiger, A.,
Lewis, O., Merullo, J.\ (2026). \emph{Reasoning Theater: Disentangling
Model Beliefs from Chain-of-Thought.} arXiv:2603.05488.

\bibitem[Chen et al.(2025)]{chen}
Chen, Y., Benton, J., Radhakrishnan, A., Uesato, J., Denison, C., et al.\
(2025). \emph{Reasoning Models Don't Always Say What They Think.}
arXiv:2505.05410.

\bibitem[Karvonen et al.(2025)]{oracles}
Karvonen, A., Chua, J., Dumas, C., Fraser-Taliente, K., Kantamneni, S.,
Minder, J., Ong, E., Sen Sharma, A., Wen, D., Evans, O., Marks, S.\
(2025). \emph{Activation Oracles: Training and Evaluating LLMs as
General-Purpose Activation Explainers.} arXiv:2512.15674.

\bibitem[Cox et al.(2026)]{cox}
Cox, K., Kianersi, D., Garriga-Alonso, A.\ (2026). \emph{Decoding
Answers Before Chain-of-Thought: Evidence from Pre-CoT Probes and
Activation Steering.} arXiv:2603.01437.

\bibitem[Esakkiraja et al.(2026)]{esakkiraja}
Esakkiraja, E., Rajeswar, S., Akhiyarov, D., Venkatesaramani, R.\ (2026).
\emph{Therefore I am. I Think.} arXiv:2604.01202.

\bibitem[Lanham et al.(2023)]{lanham}
Lanham, T., Chen, A., Radhakrishnan, A., et al.\ (2023).
\emph{Measuring Faithfulness in Chain-of-Thought Reasoning.}
arXiv:2307.13702.

\bibitem[Lindsey et al.(2025)]{lindsey}
Lindsey, J., Gurnee, W., Ameisen, E., et al.\ (2025).
\emph{On the Biology of a Large Language Model.}
Transformer Circuits Thread.
\url{https://transformer-circuits.pub/2025/attribution-graphs/biology.html}

\bibitem[Mirtaheri and Belkin(2026)]{mirtaheri}
Mirtaheri, P., Belkin, M.\ (2026). \emph{Catching rationalization in the
act: detecting motivated reasoning before and after CoT via activation
probing.} arXiv:2603.17199.

\bibitem[Pan et al.(2024)]{latentqa}
Pan, A., Chen, L., Steinhardt, J.\ (2024). \emph{LatentQA: Teaching LLMs
to Decode Activations Into Natural Language.} arXiv:2412.08686.

\bibitem[Scalena et al.(2026)]{scalena}
Scalena, D., Candussio, S., Bortolussi, L., Fersini, E., Nissim, M.,
Sarti, G.\ (2026). \emph{Beyond the Commitment Boundary: Probing
Epiphenomenal Chain-of-Thought in Large Reasoning Models.}
arXiv:2606.13603.

\bibitem[Turpin et al.(2023)]{turpin}
Turpin, M., Michael, J., Perez, E., Bowman, S.~R.\ (2023).
\emph{Language Models Don't Always Say What They Think: Unfaithful
Explanations in Chain-of-Thought Prompting.} arXiv:2305.04388.

\bibitem[Yang et al.(2025)]{qwen3}
Yang, A., Li, A., Yang, B., et al.\ (2025). \emph{Qwen3 Technical Report.}
arXiv:2505.09388.

\end{thebibliography}
\end{document}